\documentclass[11pt]{article}
\usepackage{dialogue2021}

\usepackage{ifxetex}
\ifxetex
    \usepackage{fontspec}
    \setromanfont{Times New Roman}
\else
    \usepackage[utf8]{inputenc}
    \usepackage[T2A,T1]{fontenc}
    \usepackage{times}
    \usepackage{latexsym}

\fi

\usepackage[russian,british]{babel}
\usepackage{url}
\usepackage{pgf}

\usepackage{covington} 

\dialogfinalcopy 
\graphicspath{ {./images/} }

\title{Unreasonable Effectiveness of Rule-Based Heuristics in Solving Russian SuperGLUE Tasks}

\author{Iazykova Tatyana \\
  HSE University \\
  Moscow, Russia \\
  {\tt tvyazykova@edu.hse.ru} \\
  \\
  \textbf{Kapelyushnik Denis} \\
  HSE University \\
  Moscow, Russia \\
  {\tt dmkapelyushnik@edu.hse.ru} \\\And
  Bystrova Olga \\
  HSE University / Sberbank \\
  Moscow, Russia \\
  {\tt ovbystrova@edu.hse.ru} \\
  \\
  \textbf{Kutuzov Andrey} \\
  University of Oslo \\
  Oslo, Norway \\
  {\tt andreku@ifi.uio.no}
  }

\date{}

\begin{document}

\maketitle
\bigskip
\bigskip
\bigskip
\bigskip
\begin{abstract}
Leaderboards like SuperGLUE are seen as important incentives for active development of NLP, since they provide standard benchmarks for fair comparison of modern language models. They have driven the world's best engineering teams as well as their resources to collaborate and solve a set of tasks for general language understanding. Their performance scores are often claimed to be close to or even higher than the human performance.
These results encouraged more thorough analysis of whether the benchmark datasets featured any statistical cues that machine learning based language models can exploit. For English datasets, it was shown that they often contain annotation artifacts. This allows solving certain tasks with very simple rules and achieving competitive rankings. 
  
  In this paper, a similar analysis was done for the Russian SuperGLUE (RSG), a recently published benchmark set and leaderboard for Russian natural language understanding. We show that its test datasets are vulnerable to shallow heuristics. Often approaches based on simple rules outperform or come close to the results of the notorious pre-trained language models like GPT-3 or BERT. It is likely (as the simplest explanation) that a significant part of the SOTA models performance in the RSG leaderboard is due to exploiting these shallow heuristics and that has nothing in common with real language understanding. We provide a set of recommendations on how to improve these datasets, making the RSG leaderboard even more representative of the real progress in Russian NLU.
  
  
  \textbf{DOI:} 10.28995/2075-7182-2021-20-XX-XX
\end{abstract}

\selectlanguage{british}

\section{Introduction}
\label{sec:introduction}
These days many researchers are coming to a dreadful realisation that we are not that much advanced in natural language understanding (NLU) as we used to think. Huge Transformer-based models are crowning the SuperGLUE leaderboard \cite{NEURIPS2019_4496bf24}, yet one should not trust these shining examples so fast. It has been shown in \cite{mccoy-etal-2019-right} that actually these models are exploiting statistical patterns related to the lack of diversity in data or class imbalances to demonstrate amazing performance without looking deeper and truly emulating natural language understanding. The danger of having such statistical cues is not in their mere presence. The core of the problem is that they are inherent to particular datasets only and therefore are hardly applicable to the language itself. It means that the systems do not really `understand' natural language: instead, they are utilising statistical cues that are typical to these specific datasets, so the whole process comes down to simple pattern matching. Modern language models are trained on the amount of data no native speaker will hardly ever see \cite{linzen-2020-accelerate}. But are they really as superior as we believe them to be, and is it even necessary to do genuine NLU to solve the test sets at this level of performance?

The issue became even more relevant now that the SuperGLUE benchmark, that was initially created for English, was adopted for Russian in the form of Russian SuperGLUE (RSG) benchmark and the corresponding leaderboard \cite{shavrina-etal-2020-russiansuperglue}. In this paper, we study the possibility to achieve results comparable to ones in the leaderboard without using any machine learning algorithms. We manually examined the datasets in order to find statistical regularities. As a result, we came up with a list of simple rule-based heuristics (for instance, label instances as `entailment' if they contain the word \foreignlanguage{russian}{`был'} `\textit{was}'). We do not have direct proofs that machine learning based models also make use of these shallow heuristics in the case of RSG. But we do know that this was confirmed to be true for the English SuperGLUE \cite{rogers-etal-2020-primer}, and we know that deep neural nets are extremely efficient in capturing regularities useful for their objective function. Following the Occam's Razor, we argue that finding and exploiting shallow statistical cues (not necessarily the ones we found manually) is much more plausible explanation for the observed performance of pre-trained language models than the assumption that they `understand' Russian discourse.

Moreover, we evaluated a set of trivial baselines, such as random choice, majority class and random balanced choice. The goal was to compare state-of-the-art (SOTA) results against those and to see whether cutting-edge deep learning architectures (GPT-3, BERT, etc) significantly outperform them. As we found out, this is not always the case. 

\subsection{Contributions}

The contributions of this work can be formulated as follows:
\begin{enumerate}
    \item We introduced a set of simple rule-based heuristics applicable to various datasets of Russian SuperGLUE benchmark\footnote{\url{https://github.com/tatiana-iazykova/2020_HACK_RUSSIANSUPERGLUE}}, and evaluated their performance on the test data.
    \item We evaluated the performance of even more trivial baselines (random choice, majority class, etc) on the Russian SuperGLUE tasks, to establish a lower boundary for language models' performance.
    \item A number of suggestions, spotted annotation errors and generally problematic or controversial cases are given for the authors of the Russian SuperGlUE benchmark, for further improvement.
\end{enumerate}

\section{Previous work}
\label{sec:previous_work}

Leaderboards provide the NLP community with tools to evaluate language models. This competition ensures a fair ground for comparison as the models are required to solve the same tasks on a single independently curated set of data. For example, the GLUE leaderboard \cite{wang-etal-2018-glue} was initially designed for English and consists of several diverse natural language understanding tasks and a diagnostic dataset with openly available labels to evaluate models. 

By March 2021, the situation with the GLUE dataset is the following: $14$ different models hold a higher ranking than the human performance which is equal to $87.1$ \cite{nangia-bowman-2019-human}). The knowledge about language is considered as key to solve the GLUE or any NLU tasks, yet when the SOTA approach \cite{zhang-etal-2019-ernie} (as of now) exceeded human performance by $3.8$, the creators of this model hypothesised that it was not necessary for those specific datasets. With 14 other models outperforming the human level as well, it has soon become clear that the benchmark itself is no longer able to provide a challenging evaluation system. As a result, the authors of GLUE designed SuperGLUE \cite{NEURIPS2019_4496bf24} for a more representative analysis of the current progress in NLU. To track this progress for other languages, other researchers created language-specific benchmarks similar to GLUE and SuperGLUE, e.g. Russian SuperGLUE \cite{shavrina-etal-2020-russiansuperglue} explored in this paper or CLUE \cite{xu-etal-2020-clue} for the Chinese language.

Although the SuperGLUE benchmark is more recent, its current SOTA score of $90.3$ \cite{he2021deberta} also managed to surpass the human performance \cite{NEURIPS2019_4496bf24} by $0.5$. As these competitions attract the world's best engineering teams with almost unlimited resources, models like T5 \cite{2020t5}, GPT-3 \cite{NEURIPS2020_1457c0d6}, BERT \cite{devlin-etal-2019-bert} and its optimized versions like RoBERTa \cite{liu2019roberta} usually hold top rankings and yet their performance scores differ by a mere fraction. These models prove their reputation by achieving scores that are very close to or even higher than human benchmark, and this is where some room for criticism appears. 

Such complex models require considerable resources, raising questions about their general utility \cite{ethayarajh-jurafsky-2020-utility}. Indeed, for the majority of us the size and efficiency of a model is as important as the performance scores, and some trade-off has to be allowed. Through such discussions, e. g. \cite{Rogers_2019_leaderboards}, the NLP community attempts to increase the transparency of benchmarks. Fortunately, the leaderboards are open for changes and new functionality. For example, the MOROCCO project has been recently launched to evaluate Russian SuperGLUE models in two additional dimensions: inference speed and GPU RAM usage\footnote{\url{https://russiansuperglue.com/performance/}}. 

Although these issues are important, another question --- probably a deeper one --- is raised by how exactly large-scale language models are `solving' certain NLU tasks. For example, BERT has skyrocketed the performance in many NLP tasks for English, yet if we take a closer look into its `language skills', we might be disappointed \cite{rogers-etal-2020-primer}. It appears that BERT never misses an opportunity to use shallow heuristics while solving tasks on natural language inference \cite{mccoy-etal-2019-right, zellers-etal-2019-hellaswag, Jin_Jin_Zhou_Szolovits_2020}, reading comprehension \cite{Rogers2020GettingCT, Si2019WhatDB,  Sugawara_Stenetorp_Inui_Aizawa_2020, Yogatama2019LearningAE},
argument reasoning comprehension \cite{niven-kao-2019-probing} and text classification \cite{Jin_Jin_Zhou_Szolovits_2020}. 

The above-mentioned analysis is mostly English-centred, and we are truly grateful to the creators of the Russian SuperGLUE \cite{shavrina-etal-2020-russiansuperglue}, since it is now possible to have a fair ground for comparing Russian NLU models. It is the first standardized set of diverse NLU benchmarks for Russian.
Some of the instances for its datasets were translated from the corresponding tasks in the SuperGLUE, while the others were collected by the RSG authors from scratch \cite{fenogenova-etal-2020-read}. 

In this paper, we explore all the datasets thoroughly to test their vulnerability to shallow heuristics. The results are compared to other approaches represented in the Russian SuperGLUE leaderboard. It should be noted that the RSG has been created very recently, and the human performance of $0.811$ is still at the top of the leaderboard. As of early May 2021, the highest score of $0.679$ was achieved by an ensemble of Transformer models.

\section{Methodology} 
\label{section:methodology}

The Russian SuperGLUE benchmark consists of 9 datasets or tasks, that follow the GLUE and SuperGLUE methodology. Each task is designed to evaluate if a model or an approach can solve problems with the help of logic, common sense and reasoning. Data is split into training, validation and test samples. The true labels of the test set are not openly available and to evaluate a system on the test set, it is necessary to submit the predictions to the leaderboard. Currently there are two versions of Russian SuperGLUE present, namely 1.0 and 1.1; our research was based on the latest 1.1 version.

Our general approach was to identify shallow heuristics and design rule-based functions that would surpass the results achieved by the trivial baselines (majority class, random choice and random balanced choice) and potentially approach SOTA scores. Being native Russian speakers, we invested our efforts into manual exploration of each dataset. Additionally, ELI5\footnote{\url{https://github.com/eli5-org/eli5}}, a tool to debug machine learning classifiers and explain their predictions, was applied to some of the tasks. It was used to check if any tokens are more specific to one of the classes in the dataset. Moreover, whenever the lemmatisation was needed, \texttt{pymorphy2} morphological analyzer \cite{pymorphy2} was used.    

As the datasets differ significantly, there was no intention to identify a single heuristic to solve them all: we analyzed them separately. Heuristics found in the training sets were applied to the validation sets to get an idea of their performance. All of the heuristics that were proved to work on training and validation sets were combined into functions with a set of if-else statements. To determine the order of these statements, we tested different sequences empirically and chose the ones with the higher performance scores. 

Finally, these rule-based functions were applied to the relevant test sets. To handle examples that did not trigger any of the heuristics, three aforementioned baseline methods were used to predict the label. All the predictions were grouped by their baseline function and submitted to the leaderboard to receive scores for each dataset individually as well as the total score per submission. The results are shown in the Table~\ref{table: Consoludated_data} in the section~\ref{discussion}. Below we first describe task-specific heuristics in more detail.

\subsection{Linguistic Diagnostic for Russian (LiDiRus)}

Inspired by \cite{ettinger-etal-2017-towards}, the authors of the original SuperGLUE benchmark included a small curated test dataset called AX-b for the analysis of the models' overall performance. It was `provided not as a benchmark, but as a tool for error analysis, qualitative model comparison, and development of adversarial examples' \cite{wang-etal-2018-glue}.
LiDiRus is a Russian version of this dataset, where each sentence was translated from English into Russian with the help of `professional translators and linguists to ensure that the desired linguistic phenomena remain' \cite{shavrina-etal-2020-russiansuperglue}.

\begin{flushleft}
Example\footnote{Additionally, the dataset contains fields called 'knowledge', `lexical-semantics', `logic', `predicate-argument-structure' for diagnostic purposes, however they were not used to design our heuristics-based approach.}:

`sentence1': \foreignlanguage{russian}{`Кошка сидела на коврике.'} (`The cat sat on the mat.'),

`sentence2': \foreignlanguage{russian}{`Кошка не сидела на коврике.'} (`The cat did not sit on the mat.'),

`label': `not\_entailment'
\end{flushleft}

To solve an example above, one is to predict whether there is any entailment between two sentences or not.

We identified a set of heuristics for this dataset. They are grouped in Table~\ref{table:LiDiRus_heuristics}, which also demonstrates how many samples in the validation set were covered by each heuristic and the percentage of their correct predictions. Only  basic split on white-space is applied for pre-processing sentences for all heuristics but one. `All lemmas in sentences 1 and 2 overlap' required lemmatisation first.
As the dataset does not assume any training and validation samples, the corresponding parts of the Textual Entailment Recognition for Russian (TERRa) dataset from the same RSG benchmark were used to make predictions to calculate the class distribution for the majority class and random weighted baseline functions if the utterances did not trigger the use of any heuristics. TERRa's class distribution differs from LiDiRus\footnote{The labels are distributed in the following proportions: 58.4\% not\_entailment, 41.6\% entailment for LiDiRus vs. 49.15\% not\_entailment, 50.85\% entailment for TERRa} but maintains the same dataset organisation.

\begin{table}[t]
\centering
    \begin{tabular}{|c|p{0.45\textwidth}|c|c|c|}
        \hline
        & \textbf{Heuristic} & \textbf{Target label} & \textbf{Coverage} & \textbf{Correct} \\
        \hline
        \textbf{1} & Number of tokens in sentence 1 differs from sentence 2 by more than 10 & not\_entailment & 24.3\%  & 65.2\% \\
        \hline
        \textbf{2} & Sentences 1 and 2 differ by two commas & not\_entailment & 27.3\% & 64.1\% \\
        \hline
        \textbf{3} & Sentences 1 and 2 differ by two words & not\_entailment & 16\% & 66.6\% \\
        \hline
        \textbf{4} & The presence of \foreignlanguage{russian}{`и', `не', `что', `никогда', `вовсе', `это'} (`and', `not', `that', `never', `at all', `this') in only one of the two sentences & not\_entailment & 29.3\% & 66.3\% \\
        \hline
        \textbf{5} & Vocabularies of two sentences overlap by 100\% (lemmatised data) & entailment & 4\% & 64.4\% \\
        \hline
        \textbf{6} & \foreignlanguage{russian}{`Чтобы', `будет', `от', `он'} (`in order to', `will', `from', `he') occur in both sentences & entailment & 11.6\% & 57\% \\
        \hline
    \end{tabular}

    \caption{LiDiRus: identified heuristics with their coverage of the validation set and the percentage of correct predictions}
    \label{table:LiDiRus_heuristics}
\end{table}
   
The performance of the aforementioned heuristics (as well as heuristics for other RSG tasks) is consolidated into Table \ref{table: Consoludated_data} which can be found in section~\ref{discussion}. It provides SOTA scores for each task as of May 2021, performance scores of the baseline functions, as well as the results for heuristics-based approach supported by one of the three baseline functions. The evaluation metric used for LiDiRus is Matthews correlation coefficient \cite{MATTHEWS1975442}. The authors of the original benchmark for English suggested this metric, as it can be applied to unbalanced binary classification problems and its values range from -1 to 1, with 0 being the performance of uninformed guessing \cite{wang-etal-2018-glue}. 

As it is a diagnostic dataset, the SOTA approach is hardly applicable to it, though the fact that there is a small performance gap between heuristics and other models deserves to be mentioned. It supports the hypothesis that shallow heuristics might play a significant part in the results of the approaches which apply pre-trained language models to solve NLP tasks.

\subsection{Russian Commitment Bank (RCB)}

Russian Commitment Bank is a Natural Language Inference task dataset that consists of naturally occurring discourses where the task is to predict the relation of one phrase (hypothesis) to the given text (premise), where the options are entailment, contradiction and neutral. 

\begin{flushleft}
\begin{itemize}
\item \textbf{entailment} --- the text of the hypothesis can be deduced or is clear from the premise (Example: premise: \emph{\foreignlanguage{russian}{`Готовность станций в настоящее время достаточно высокая, - сказал он.'} (`He said that the stations were almost ready at that time.')}, hypothesis: \emph{\foreignlanguage{russian}{`Готовность станций достаточно высокая.'}(`The stations are almost ready.')}; 

\item \textbf{contradiction} --- the text of the hypothesis lays in clear opposition to what was said in the premise (Example: premise: \emph{\foreignlanguage{russian}{`Перебрасываясь словечками, они скользят глазами по моему городу. Как они смеют смотреть, будто что-то понимают?'}(`Exchanging phrases, their eyes passed over my town. How dare they look as if they understand something.')}, hypothesis: \emph{\foreignlanguage{russian}{`Они что-то понимают.'} (`They understand something.'))};

\item \textbf{neutral} --- the relation between the hypothesis and the premise is hard to establish (Example: premise: \emph{\foreignlanguage{russian}{`За происходящим наблюдал очень толстый мужчина. Я заметил в его глазах ревность. Мне показалось, что это был местный спортивный босс.'} (`A very fat man was watching the situation. I noticed jealousy in his eyes. It seemed to be that it was local sports boss.')}, hypothesis: \emph{\foreignlanguage{russian}{`Это был местный спортивный босс.'} (`It was local sports boss.'))}. 
\end{itemize}

\end{flushleft}

The training data is distributed unequally in this dataset (46.3\% --- neutral, 35.4\% --- entailment, 18.3\% --- contradiction for train data; 52.7\% --- neutral, 33.6\% --- entailment, 13.6\% --- contradiction for validation data). This imbalance can potentially lead to a substantial bias towards a certain class for the large pre-trained language models. The model can simply predict the majority class and still achieve a rather good result, though it by any means would not be natural language understanding. 

\begin{figure}
\includegraphics[scale=0.5,keepaspectratio]{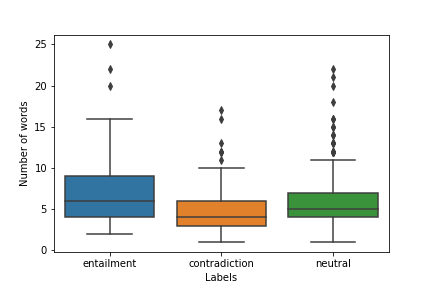}
\includegraphics[scale=0.5,keepaspectratio]{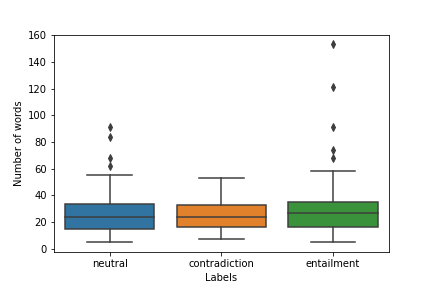}
\caption{RCB: interplay between the number of words and the label. Left: the number of words in the hypothesis. Right: the number of words in the premise.}
\label{table:RCB_numberofwords}
\end{figure}

The number of instances in the training data is 438, in the validation --- 220 and in the test --- 438. Two metrics are used to evaluate the model's performance on solving this task: Accuracy and F1, as is the case with the corresponding Commitment Bank task in the original SuperGLUE \cite{wang-etal-2018-glue}. According to the authors of the SuperGLUE, the imbalanced nature of the dataset (relatively fewer neutral examples in the English version and significantly more neutral instances in the Russian SuperGLUE) was the reason for them using two metrics, where they used macro-F1 for multi-class problems.

One of the heuristics (you can see its performance in Table \ref{table:RCB_heuristics}) used for solving this dataset utilised the correlation between the label and the number of words in the hypothesis or the premise. 
Figure \ref{table:RCB_numberofwords} illustrates this phenomenon. 
From the left plot, it is clear that instances with 5-7 words in the hypothesis would more likely have the `neutral' label (median for it is 5) and instances with less than 5 words in the hypothesis would more likely have the `contradiction' label (median for it is 4). 
As for the right plot, one may notice that instances with more than 30 words in them would likely belong to the `entailment' category (median for the `entailment' is 27).

\begin{table}
\centering
\begin{tabular}{|c|p{0.45\textwidth}|c|c|c|}
\hline
 &\textbf{Heuristic} & \textbf{Target} \textbf{label} & \textbf{Coverage} & \textbf{Correct} \\
\hline
1 & The hypothesis is a sub-string of the premise & entailment & 26\% & 40\% \\
\hline
2 & 75\% intersection of the hypothesis and premise's vocabularies (lemmatised data)& entailment & 5\% & 45\% \\
\hline
3 & The presence of \foreignlanguage{russian}{`признать'} (`admit') (lemmatised data)& entailment & 6\% & 36\% \\
\hline
4 & The presence of \foreignlanguage{russian}{`подозревать, cчитать, говорить, думать, надеяться, понять, уверять'} (`suspect, consider, say, think, hope, assure, realise') (lemmatised data)& neutral & 6\% & 36\% \\
\hline
5 & Hypothesis $> 5$ words & contradiction& 41\% & 23\%\\
\hline   
6 & $4 <$ hypothesis $< 8$ words & neutral & 34\% & 70\%\\
\hline
7 & More than 30 words in the premise & entailment & 35\% & 39\%\\
\hline 
\end{tabular}
\caption{RCB: identified heuristics with their coverage of the validation set and the percentage of correct predictions}
\label{table:RCB_heuristics}
\end{table}

As we can see from the Table \ref{table:RCB_heuristics}, the heuristics do not cover all the data, leaving some answers to be predicted with the help of three baselines (majority class, random choice, random balanced choice). The results achieved with the help of the heuristics were comparable with results of large pre-trained language models in the RSG leaderboard, which are given in the section \ref{discussion} of this paper.

\subsection{Choice of Plausible Alternatives for Russian language (PARus)}
To evaluate progress in open-domain common sense casual reasoning, the authors of Russian SuperGLUE provided the Choice of Plausible Alternatives for Russian language (PARus) dataset. It is based on the English COPA \cite{roemmele2011choice}. A typical task in PARus consists of a premise and two alternatives, where the goal is to select the alternative that has a causal or effect relation with the premise. 

\begin{flushleft}
Example:
{
`premise': {\foreignlanguage{russian}{`Гости вечеринки прятались за диваном.'}}, 
(`The guests were hiding behind the couch')

`choice1': {\foreignlanguage{russian}{`Это была вечеринка-сюрприз.'}}, (`It was a surprise party')

`choice2':{\foreignlanguage{russian}{`Это был день рождения.'}}, (`It was a birthday party')

'question': 'cause',

'label': 0
}
\end{flushleft}

In the example above, we have the premise and two reasons for why this situation could happen in the first place. Our task is to choose one of them based on their semantic meaning and the `question' field (this field can take either `effect', or `cause' values). Here it is obvious for humans that the situation in the premise can happen probably because of the first alternative, as it is more probable that guests would hide behind the couch because they want to throw a surprise party for someone. 

There are 400 samples in the train dataset and 100 in the validation set. Since there is no semantics behind the labels, the difference between label distribution in the training and validation data should be considered irrelevant. Also because of this lack of label meaning, it was challenging to find linguistic heuristics to solve this task. All textual data was lemmatised to get better results. The heuristics used for tackling this task are shown in Table \ref{table:PARus_heuristics}.

The heuristics check whether one of the choices has more shared lemmas with the premise than the others, and if so, then this choice should be taken as an answer. If the vocabulary overlap was the same for all choices, one of the baseline functions was applied. If one of choices had more words than the other, then this choice was taken as an answer. 

\begin{table}[ht]
\centering
\begin{tabular}{|c|p{0.65\textwidth}|c|c|}
\hline
&\textbf{Heuristic} & \textbf{Coverage} & \textbf{Correct} \\
\hline
1 & If one of choices has more shared lemmas with the premise than the others, it is taken as an answer (lemmatised data)& 22\% & 64\% \\
\hline
2 & If one of choices has more words than the others, then this choice should be taken as an answer (lemmatised data)& 59\% & 52\% \\
\hline 
3 & The combination of these two heuristics (lemmatised data)& 66\% & 52\% \\
\hline
\end{tabular}
\caption{PARus: identified heuristics with their coverage of the validation set and the percentage of correct predictions}
\label{table:PARus_heuristics}
\end{table}

As we can see from Table \ref{table:PARus_heuristics}, these heuristics cover less than 70\% of the data, therefore many answers still depend on one of three baselines. Overall results are presented in the Table \ref{table: Consoludated_data} in section \ref{discussion}. The maximal accuracy score was $0.516$. To achieve SOTA performance, we probably need more complex algorithms. It proves that this task fulfills its goal and we do need to learn some open-domain common sense casual reasoning to solve such tasks. 

\subsection{Russian Multi-Sentence Reading Comprehension (MuSeRC)}

The MuSeRC dataset is collected for the reading comprehension task. It contains more than 900 paragraphs across 5 different domains: elementary school texts, news, fiction stories, fairy tales, and summaries of TV series and books \cite{fenogenova-etal-2020-read}. Samples were collected based on the following criteria: 
\begin{enumerate}
    \item the passage length is less than 1.5K characters;
    \item the passage contains named entities;
    \item if the passage contains only one named entity, then it must have one or more co-reference relations.
\end{enumerate}

Furthermore, the authors of the dataset ensured correct sentence splitting and used these sentences in a crowd-sourcing effort at the Yandex.Toloka platform. In it, humans were asked to generate questions, a set of answers for each of them and to check that answering a question requires consulting with more than one sentence in the text. The answer can be either True or False, so all the answers are either correct or incorrect with no in-between. The number of correct answers varies and each question/answer pair is treated individually\footnote{The average number of questions is approximately 20. The labels are distributed in the following proportions: 55\% false and 45\% true for the training set vs. 55.6\% false and 44.4\% true for the validation set.}.

\begin{flushleft}
Example\footnote{The example was shortened due to the page limit. The original sample contains more sentences (each of them was enumerated by the authors of the dataset on purpose) in a text column and a higher number of questions.}:

`text': \foreignlanguage{russian}{`(11) Напомним, что днем ранее российские биатлонистки выиграли свою эстафету. (12) В составе сборной России выступали Анна Богалий-Титовец, Анна Булыгина, Ольга Медведцева и Светлана Слепцова. (13) Они опередили своих основных соперниц - немок - всего на 0,3 секунды.'}, (`(11) The day before, the Russia women's national biathlon team won their relay competition. (12) The lineup was Anna Bogaliy-Titovets, Anna Bulygina, Olga Medvedtseva and Svetlana Sleptsova. (13) They were ahead of their main rivals, the Germany women's national team, by only 0.3 seconds.') 

`question': \foreignlanguage{russian}{`На сколько секунд женская команда опередила своих соперниц?'}, (`How many seconds were the women's team ahead of their rivals?'),

`answers': 

`text': \foreignlanguage{russian}{`Всего на 0,3 секунды.'}, (`Only 0,3 seconds.'), `label': True
            
`text': \foreignlanguage{russian}{`На 0,3 секунды.'}, (`0,3 seconds.'), `label": True

`text': \foreignlanguage{russian}{`На секунду.'}, (`One second.'), `label': False

`text': \foreignlanguage{russian}{`На 0.5 секунд.'}, (`0,5 seconds.'),`label': False
\end{flushleft}

To solve the example above, one has to use information from multiple sentences in the `text' field, namely \foreignlanguage{russian}{`российские биатлонистки'} (`the Russian women's national biathlon team') from sentence 11 and \foreignlanguage{russian}{`опередили своих основных соперниц - немок - всего на 0,3 секунды'} (`were ahead of their main rivals, the Germany women's national team, by only 0.3 seconds') from sentence 13.

A set of heuristics identified for this dataset is grouped in Table \ref{table:MuSeRC_heuristics}. 

\begin{table}[htbp]
\centering
    \begin{tabular}{|c|p{0.45\textwidth}|c|c|c|}
        \hline
        & \textbf{Heuristic} & \textbf{Target label} & \textbf{Coverage} & \textbf{Correct} \\
        \hline
        1 & All lemmas from the answer occur in the text (lemmatised data)& True & 39.2\%  & 58.8\% \\
        \hline
        2 & The answer is longer than 11 tokens & True & 10.3\%  & 72.3\% \\
        \hline
        3 & More than 6 overlapping lemmas between the answer and the text (lemmatised data)& True & 18.9\%  & 73.9\% \\
        \hline
        4 & No overlapping lemmas between the answer and the text (lemmatised data)& False & 9.9\%  & 89.1\% \\
        \hline
        5 & The answer is shorter than 4 tokens & False & 46.4\%  & 64.9\% \\
        \hline
        6 & One overlapping lemma between the answer and the text (lemmatised data)& False & 18.6\% & 69.3\% \\
        \hline
    \end{tabular}

    \caption{MuSeRC: identified heuristics with their coverage of the validation set and the percentage of correct predictions}
    \label{table:MuSeRC_heuristics}
\end{table}

While predicting, the if-else statements dealt with the `True' label first, as it is less frequent in the data. The function yields the intended label as long as at least one of the heuristics gets triggered. After that, the opposite set of heuristics is applied. 

The overall performance is given in Table~\ref{table: Consoludated_data} which can be found in section~\ref{discussion}. To provide the evaluation metrics, the dataset authors roughly followed the evaluation procedure by \cite{DBLP:journals/corr/abs-1810-12885, Khashabi2018LookingBT}. Since each answer-option can be assessed independently, F1-averaged (F1a) is applied to evaluate binary decisions over all the answer options in the dataset. It is a harmonic mean of precision and recall per question. Exact Match (EM) is the exact match per each instance, i.e. each set of predictions should be the same as of the answers \cite{fenogenova-etal-2020-read}.

We were not able to reach neither the SOTA score nor the human performance, although the obtained results are on par with some of those produced by large pre-trained language models. In fact, at the time of submission, our heuristics-based approach combined with the majority class baseline function achieved higher performance scores for this task than Multilingual Bert and RuGPT3Small\footnote{\url{https://huggingface.co/sberbank-ai/rugpt3small_based_on_gpt2}}.

\subsection{Textual Entailment Recognition for Russian (TERRa)}

Textual Entailment Recognition is another dataset dedicated to the Natural Language Inference task. This task requires to recognise, given two text fragments, whether the meaning of one text is entailed (can be inferred) from the other text \cite{shavrina-etal-2020-russiansuperglue}. This task is similar to the RCB, yet in TERRa there are only two categories (entailment/not\_entailment) instead of three.
The number of instances in the training data is 2 616, in the validation --- 307 and in the test --- 3 198. 

\begin{flushleft}
Examples:
\textbf{entailment}

premise: \foreignlanguage{russian}{`Женщину доставили в больницу, за ее жизнь сейчас борются врачи.'}(`A woman was brought to the hospital, doctors are continuing to work on her right now.')

hypothesis: \foreignlanguage{russian}{`Женщину спасают врачи.'} (`The woman is being treated by the doctors')

\textbf{not\_entailment}

premise: \foreignlanguage{russian}{`О случившемся она заявила в полицию. Когда супруг вернулся из командировки и обо всем узнал, то принял решение с женой расстаться. Официально они не развелись, но живут отдельно.'} (`She reported about what had happened to the police. When her spouse returned from the business trip and learned everything, he decided to broke up with his wife. They are not officially divorced but they live separately')

hypothesis: \foreignlanguage{russian}{`Супруги живут вместе.'}(`Spouses live together')

\end{flushleft}

\begin{figure}[ht]
\begin{center}
\includegraphics[scale=0.5]{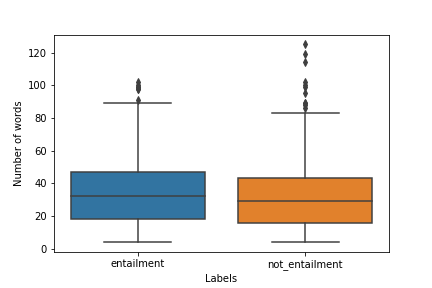}

\end{center}
\caption{TERRa: interplay between the number of words and the label in the premise}
\label{fig:TERRa_numberofwords}
\end{figure}

Similar to the RCB, one of the heuristics (Table \ref{table:TERRa_heuristics}, heuristics 6 and 7) used in solving this dataset utilised the interplay between the label and the number of words. Unlike with the RCB, in this dataset it was possible to find such relations only between the label and the number of words in the premise. 
Instances with less than 29 words would more likely have the label `not\_entailment' (median number of words for not\_entailment is 29) whereas if the number of words was more than 32, then the label is likely`entailment' (median number of words for entailment is 32).  
Figure \ref{fig:TERRa_numberofwords} illustrates this phenomenon for the training data.

Another heuristic (8 in Table~\ref{table:TERRa_heuristics}) for TERRa dealt with the presence of specific words, namely \foreignlanguage{russian}{`только'} (`only') and \foreignlanguage{russian}{`мужчина'} (`man') in the hypothesis. It was possible to find a rather noticeable correlation between their presence and the label. You can find the illustrating examples in the Appendix~\ref{examples}.

\begin{table}[ht]
\centering
\begin{tabular}{|c|p{0.45\textwidth}|c|c|c|}
\hline
&\textbf{Heuristic} & \textbf{Target label} & \textbf{Coverage} & \textbf{Correct} \\
\hline
1 & The hypothesis is a sub-string of the premise & entailment & 1\% & 50\% \\
\hline
2 & Vocabularies of the hypothesis and the premise overlap by 33\%  (lemmatised data)& not\_entailment & 11\% & 69\% \\
\hline
3 & Vocabularies of the hypothesis and the premise overlap by 75\% (lemmatised data)& entailment & 9\% & 52\% \\
\hline
4 & Vocabularies of the hypothesis and the premise overlap by 66\% (lemmatised data)& entailment & 9\% & 56\% \\
\hline
5 & Vocabularies of the hypothesis and the premise overlap by 100\% (lemmatised data)& entailment & 14\% & 65\% \\
\hline
6 & Less that 29 words in the premise & not\_entailment & 45\% & 58\%\\
\hline
7 & More than 32 words in the premise & entailment & 45\% & 60\%\\
\hline
8 & The presence of \foreignlanguage{russian}{`только', `мужчина'} (`only', `man') (lemmatised data)& not\_entailment & 21\% & 66\%\\
\hline     
\end{tabular}
\caption{TERRa: identified heuristics with their coverage of the validation set and the percentage of correct predictions}
\label{table:TERRa_heuristics}
\end{table}

As we can see from the Table \ref{table:TERRa_heuristics}, the heuristics do not cover all the data, leaving some answers to be predicted with the help of the three trivial baselines. However, the results achieved with the help of the heuristics were comparable with the results of large pre-trained language models in the RSG leaderboard and even outperformed the ones by RuGPT3Medium and RuGPT3Small. 

\subsection{Russian Words in Context (RUSSE)}

Depending on its context, a word can have multiple, potentially unrelated, senses. For example, the Russian word \foreignlanguage{russian}{`лук'} ('onion'/'bow') can mean either vegetable or weapon depending on its surrounding words. The `word in context' task can be described as a binary classification problem, and the goal is to predict whether a given word has the same meaning in both given sentences.
The Russian SuperGLUE task borrows original data from the Russe Word Sense Induction and Disambiguation shared task \cite{Panchenko:18}. 

\begin{flushleft}
Example:

'word' : {\foreignlanguage{russian}{`дорожка'}}, ('road / carpet')

'sentence1' : {\foreignlanguage{russian}{`Бурые ковровые дорожки заглушали шаги'}}, ('Greyish-brown carpets drowned out steps')
  
'sentence2' : {\foreignlanguage{russian}{`Приятели решили выпить на дорожку в местном баре}}, ('Two friends decided to get drinks at the local bar before their road trip')

'label' : false
\end{flushleft}

In the example above, we have two sentences (`sentence1' and `sentence2') and our task is to decide whether the target word has the same meaning in both of them. Again, for a native Russian speaker it is obvious that the word \foreignlanguage{russian}{`дорожка'} has different meanings in two sentences. 

To find out whether the decision can be made based on simple rules, we checked whether the target word appears in the same form in both sentences. In addition, we calculated the proportion of shared tokens to all tokens in both sentences and the difference in their lengths.

\begin{table}[ht]
\centering
\begin{tabular}{|c|p{0.45\textwidth}|c|c|c|}
\hline
& \textbf{Heuristic} & \textbf{Target label} & \textbf{Coverage} & \textbf{Correct} \\
\hline
1 & Target word in the same form & True & 14\% & 58\% \\
\hline
2 & Tokens overlap by more than 10\%& True& 4\% & 50\% \\
\hline 
3 & Number of tokens in sentence 1 differs from sentence 2 by more than 6 & False & 49 \% & 65 \% \\
\hline
\end{tabular}
\caption{RUSSE: identified heuristics with their coverage of the validation set and the percentage of correct predictions}
\label{table:RUSSE_heuristics}
\end{table}

The heuristics cover about 50\% of the data and make correct predictions in about 65\% cases. According to Table \ref{table: Consoludated_data} in Section \ref{discussion}, we managed to achieve $0.595$ accuracy score.

\subsection{The Winograd Schema Challenge for Russian (RWSD)}

The original purpose of the Winograd Schema Challenge (WSC) was to serve as an alternative Turing test to evaluate an automatic system’s capacity for common sense inference \cite{10.5555/3031843.3031909}. The challenge evaluates the models' ability to identify the antecedent of the pronoun, which might be critical, for example, for translation purposes \cite{davis2016winograd}. The performance scores on the WSC for English quickly progressed from a simple guess to near-human level \cite{emami-etal-2020-analysis} after neural language models trained on massive
corpora were applied to solve this challenge.

The RWSD dataset is a Russian translation of the pronoun disambiguation task used in the SuperGLUE benchmark \cite{Morgenstern_Davis_Ortiz_2016}. RWSD maintains the same structure providing a pair or a batch of sentences that differ by one or two words:

\begin{flushleft}
Example 1: \foreignlanguage{russian}{`Кубок не помещается в коричневый чемодан, потому что он слишком \textbf{большой}.'} (`The trophy doesn't fit into the brown suitcase because it is too large.')

Example 2: \foreignlanguage{russian}{`Кубок не помещается в коричневый чемодан, потому что он слишком \textbf{маленький}.'} (`The trophy doesn't fit into the brown suitcase because it is too small.')
\end{flushleft}

There is an ambiguity in these sentences, namely \foreignlanguage{russian}{`он'} (`it') might refer to either \foreignlanguage{russian}{`кубок'} (`the trophy') or \foreignlanguage{russian}{`чемодан'} (`suitcase'). Each sentence is followed by an antecedent and a pronoun for disambiguation, which can be successfully resolved if a model assigns a `false' label to the first example for the pair of \foreignlanguage{russian}{`чемодан'}(`suitcase') and \foreignlanguage{russian}{`он'} (`it'), and if `true' is assigned to the second example for the same pair. 

The model cannot rely on the word order or the structure of a sentence, as the task is organised so that they cannot be used for the disambiguation process \cite{Morgenstern_Davis_Ortiz_2016}. Each sentence might be either `true' or `false' depending on a suggested pair of antecedents and pronouns. For example, one has to pay attention to a special word, i.e.  \foreignlanguage{russian}{`большой'} (`large') or \foreignlanguage{russian}{`маленький'} (`small') in the aforementioned sentences.

There is a clearly unequal distribution of classes in the RWSD dataset. The labels for the training and validation sets are distributed as follows: $51$\% `false' and $49$\% `true' labels for the former and $55.4$\% `false' and $44.6$\% `true' labels for the latter. However, the `false' values appear $67\%$ of the times in the \textit{test set}, which is very different from the datasets provided for training and validation.

We were not able to identify any heuristic that would surpass the performance score of predictions made by the majority class baseline (see Table \ref{table: Consoludated_data} for reference), but this misfortune carries one of our most important findings. 

Apparently, the very same approach to choose the most common value was used by many sophisticated models listed in the Russian SuperGLUE leaderboard by the time of our submission. The SOTA score, which is the score achieved by Multilingual T5, several BERT variations (trained on multilingual data and on Russian corpora only), RuGPT3Medium and RuGPT3Small, is $0.669$: the same as achieved by our majority class baseline function. While solving the task, these models allegedly opted to predict using the majority class rather than try to actually solve the Winograd Schema Challenge. Such models as Golden Transformer, RuGPT3XL few-shot and RuGPT3Large apparently made an attempt to really predict something, but their results are sub-optimal:$0.545$, $0.649$ and $0.636$ respectively, which is in fact below the $0.662$ tf-idf baseline provided by the RSG creators. The problem is similar to that with Winograd Schema Challenge (WSC) in the original SuperGLUE \cite{wang-etal-2018-glue}.

\subsection{Yes/no Question Answering Dataset for Russian (DaNetQA)}

DaNetQA is a question answering dataset for yes/no questions. Each example is a triplet of (passage, question, answer), with the title of the page as optional context \cite{shavrina-etal-2020-russiansuperglue}. The answers are encoded in a True/False formal similar to the corresponding SuperGLUE `BoolQ' dataset. As with the Russian Commitment Bank task, here we can also notice the unequal distribution of labels (Train: True --- 60.7\%, False --- 39.3\%) and the mismatch of this relation among training and validation data (Vaidation: True --- 50.2\%, False --- 49.8\%). 
The number of instances in the training data is 1 749, in the validation --- 821 and 805 in the test set.

\begin{flushleft}
Example\footnote{The example was shortened due to the page limit. The original sample contains more sentences in the passage.}:

question: \foreignlanguage{russian}{ВДНХ --- это выставочный центр?}(`Is VDNKh an exhibition centre?')

passage: \foreignlanguage{russian}{Выставка достижений народного хозяйства, в 1959—1991 годах — Выставка достижений народного хозяйства СССР , в 1992—2014 годах — Всероссийский Выставочный Центр) — выставочный комплекс в Останкинском районе Северо-Восточного административного округа города Москвы, второй по величине выставочный комплекс в городе.  <...> На территории Выставки расположено множество шедевров архитектуры — 49 объектов ВДНХ признаны памятниками культурного наследия.} (`Exhibition of Achievements of National Economy from 1959 to 1991 Exhibition of Achievements of National Economy USSR, from 1992 to 2014 All-Russia Exhibition Centre is and exhibition facility in the Ostankino district in the North-Eastern Administrative Okrug of Moscow city. <...> There are plenty of architectural masterpieces located in the Exhibition, 49 of which are recognised as cultural heritage sites.') 

label: True
\end{flushleft}

\begin{figure}[t]
\includegraphics[scale=0.5,keepaspectratio]{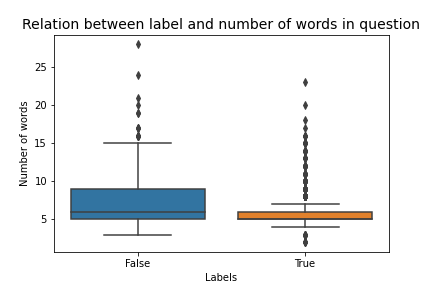}
\includegraphics[scale=0.5,keepaspectratio]{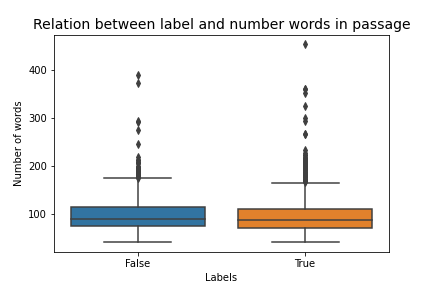}
\caption{DaNetQA: interplay between the number of words and the label}
\label{fig:DaNetQA_numberofwords}
\end{figure}

Like with the RCB and TERRa, one of the heuristics used in solving this dataset utilised the relations between the label and the number of words in questions (Table \ref{table:DaNetQA_heuristics}, heuristic 6) or passage (heuristic 7). 
One can see this phenomenon in Figure \ref{fig:DaNetQA_numberofwords}.
From the left plot it is clear that instances with more than 5 words  would more likely have the label `False' (median number of words for False is 6 in the training data). 
As for the right plot, one may notice that instances with more words in the passage have slightly more chances to belong to the `False' category (median number of words for False is 90 in the training data whereas median number of words for True is 88). 

Heuristic 3 exploits correlation between the beginning of the question and the label: if the question starts with \foreignlanguage{russian}{`входит ли'} (`does it belong to'), the label in the validation dataset is False 100\% of the time. One can find an example for this heuristic in the Appendix \ref{examples}.

\begin{table}[t]
\centering
\begin{tabular}{|c|p{0.45\textwidth}|c|c|c|}
\hline
& \textbf{Heuristic} & \textbf{Target label} & \textbf{Coverage} & \textbf{Correct} \\
\hline
1 & The question starts with \foreignlanguage{russian}{`был'} (`was/were') & True & 45\% &  58\% \\
\hline
2 & The question starts with \foreignlanguage{russian}{`есть'} (`is/are') & True & 13\% &  81\% \\
\hline
3 & The question starts with \foreignlanguage{russian}{`входит ли'} (`does it belong to') & False & 37\% &  100\% \\
\hline
4 & The question starts with \foreignlanguage{russian}{`едят ли'} (`do they eat')  & False & 2\% &  53\% \\
\hline
5 & The question starts with \foreignlanguage{russian}{`правда ли'} (`is it true') & False & 18\% &  89\% \\
\hline
6 & More than 5 words in the question & False & 46\% &  71\% \\
\hline 
7 & More than 90 words in the passage & False & 48\% &  53\% \\
\hline
\end{tabular}
\caption{DaNetQA: identified heuristics with their coverage of the validation set and the percentage of correct predictions}
\label{table:DaNetQA_heuristics}
\end{table}

As we can see from the Table \ref{table:DaNetQA_heuristics}, the heuristics do not cover all the data, leaving some answers to be predicted with the help of the three trivial baselines. However, the results achieved with the help of the heuristics were comparable with the results of large pre-trained language models in the RSG leaderboard. 

\subsection{Russian Reading Comprehension with Commonsense Reasoning (RuCOS)}
Russian reading comprehension with Commonsense reasoning (RuCoS) is a large-scale reading comprehension dataset which requires common sense reasoning. Unlike MuSeRC, the main data domain for RuCoS is news articles and there is more data for this task. Also in this task, a system is given a list of named entities from which it should choose the right answer (while in the MuSeRC, the answers do not have to be named entities at all). RuCoS consists of queries automatically generated from news articles; the answer to each query is a text span from a summarizing passage of the corresponding article.

This task is based on the English ReCoRD  benchmark \cite{DBLP:journals/corr/abs-1810-12885}. All text examples were collected from Russian media. The texts were then filtered by the IPM frequencies of the contained words and, finally, manually reviewed. 

\begin{flushleft}
Example:

{`source': `Lenta',

`passage': {

`text':
\foreignlanguage{russian}{'Мать двух мальчиков, брошенных отцом в московском аэропорту Шереметьево,
забрала их. Об этом сообщили ТАСС в пресс-службе министерства образования и науки Хабаровского края <...> Через несколько дней после того, как Виктор Гаврилов бросил своих детей в аэропорту, он явился с повинной к следователям в городе Батайске Ростовской области. <...>'}, 

('Mother of two boys abandoned by their father in Moscow airport Sheremetyevo, got them back. TASS got that information from press service of the Ministry of Education and Science of the Khabarovsk Territory. <...> In a few days after Victor Gavrilov abandoned his children in the airport, he came to the police station in Bataysk city, Rostov region. <...>')
 
`entities': [
{`start': 60, `end': 71, `text': \foreignlanguage{russian}{`Шереметьево'} (`Sheremetyevo')},

{`start': 102, `end': 106, `text': \foreignlanguage{russian}{`ТАСС'} (`TASS')},

{`start': 155, `end': 172, `text': \foreignlanguage{russian}{'Хабаровского края'} (`Khabarovsk Territory')},

{`start': 470, `end': 485, `text': \foreignlanguage{russian}{`Виктор Гаврилов'} ('Victor Gavrilov')},

{`start': 563, `end': 571, `text': \foreignlanguage{russian}{`Батайске'} (`Bataysk')},

{`start': 572, `end': 590, `text': \foreignlanguage{russian}{`Ростовской области'} (`Rostov Region')}
]
},

`qas': [
{

    `query': 
    \foreignlanguage{russian}{`26 января @placeholder бросил сыновей в возрасте пяти и семи лет в Шереметьево.'} 

(`January 26 @placeholder left their sons in the age of 5 and 7 in Sheremetyevo airport'),

`answers': [
    {`start': 470, `end': 485, `text': \foreignlanguage{russian}{`Виктор Гаврилов'} (`Victor Gavrilov')}
              ]
          }
      ]
  }
\end{flushleft}

In the example above, we have a text with several pre-defined entities (names, cities, countries etc.). Our task was to choose one of the entities that can replace the @placeholder token in the question. All questions are placed in 'qas' field.  

The heuristics applied to this task dealt with the presence of name entities in the question. The algorithm simply predicted all the entities present in the question. A modification of this approach was to sort the remaining entities based on the frequency of their appearance in the text. We tried several threshold values for this rule and finalized our choice on the following rule: all entities whose stems appeared less than three times were removed from our predictions. 

Both heuristics were applied every time we made predictions. Thus, their coverage is 100\%. We managed to outperform the tf-idf baseline with both F1 score end EM metric around $0.26$, but the SOTA results are on par with human performance score, which is $0.93/$ for F1 and $0.89$ for EM. 

\section{Discussion}
\label{discussion}

\begin{table}[htbp]
\centering
\begin{tabular}{|l|l|r|r|r|r|r|r|r|r|}
\hline
&   \textbf{Metrics} &  \textbf {Hum.} & \textbf{SOTA} &  \textbf{maj}. &  \textbf{rand.} &  \textbf{r.(b)}&  \textbf{H maj.} &  \textbf{H rand.} &  \textbf{H r.(b)} \\
\hline
\hline
LiDiRus &  M. Corr &  {0.626} & \textbf{0.231} &     0.000 &   0.024 & 0.000 &  0.147 &  0.149 &  \textbf{0.182} \\
   \hline
RCB &  Avg. F1 & {0.680} & \textbf{0.452} & 0.217 & 0.332 &  0.319 &  0.400 &  \textbf{0.401} &   \textbf{0.401} \\

&  Acc. &  {0.702} & \textbf{0.546} &  \textbf{0.484} &  0.347 & 0.374 & 0.438 & 0.436 & 0.438 \\
       \hline
PARus & Acc. &  {0.982} & \textbf{0.908} &     0.498 &   0.474 &  0.480 & 0.478 & \textbf{0.508} &  0.470 \\
     \hline
MuSeRC & F1a &  {0.806} & \textbf{0.941} & 0.000 & 0.477 & 0.450 &  \textbf{0.671} & 0.669 & 0.669 \\
& EM &  {0.420} & \textbf{0.819} & 0.000 & 0.078 & 0.071  & 0.237 & 0.195 & \textbf{0.202} \\
    \hline
TERRa & Acc. &  {0.920} & \textbf{0.871} & 0.513 & 0.503 & 0.483 &  \textbf{0.549} & 0.547 & 0.548 \\
     \hline
RUSSE &  Acc. &  {0.805} & \textbf{0.729} & 0.587 & 0.501 & 0.528 & \textbf{0.595} & 0.497 & 0.543 \\
     \hline
RWSD &  Acc. &  {0.840} & \textbf{0.669} & \textbf{0.669} &   0.487 & 0.597  & \textbf{0.669} & 0.565 & 0.604 \\
      \hline
DaNetQA & Acc. &  {0.915} & \textbf{0.917} &     0.503 &   0.494 & 0.520  &\textbf{ 0.642} & 0.629 &  0.629 \\
   \hline
RuCoS & F1 & {0.930} & \textbf{0.920} & 0.250 & 0.250 & 0.250 &  \textbf{0.260} & 0.260 &  0.260 \\
 & EM &   {0.890} & \textbf{0.924} & 0.247 &   0.247 & 0.247 &  \textbf{ 0.257} &  0.257 & 0.257 \\
     \hline
Total  &&  {0.811} & \textbf{0.679} & 0.374 &   0.372 &  0.385 &  \textbf{0.468} & 0.445 &  0.454 \\
    \hline

\end{tabular}
\caption{Performance scores at the time of submission. `Maj.' is the majority class baseline function, `rand.' --- random choice, `r.(b)' --- random balanced choice, H --- the heuristics-based approach.}
\label{table: Consoludated_data}
\end{table}

Table~\ref{table: Consoludated_data} shows the best results after applying the heuristics described above to the Russian SuperGLUE test sets.
The heuristic based approach (H) was combined with one of the trivial baseline functions. The majority value and weights for baseline functions were obtained from combined training and validation sets.
For every task, we chose heuristic (or combination of several heuristics) that led towards the best score. Even if for some tasks we are still far away from beating SOTA performance, simple baselines and heuristics can achieve relatively good results. 
For RWSD task one can achieve SOTA performance just by using the majority class baseline.

Our heuristics approach works well for the RCB task with the difference between H maj. model and SOTA being about 5\%. For TERRa, RUSSE and DaNetQA we are far from SOTA results but, still, our results are on the same level with RuBERT \cite{kuratov2019adaptation} and GPT models from the leaderboard. 

Yet for several tasks, the heuristics approach did not work as well. RuCoS, MuSeRC and PARus proved that it is not enough to use dataset-specific statistical cues to solve them, so for these three tasks it seems that the large pre-trained language models really pick up some peculiarities of Russian language.

Since our approaches can be divided into two groups (trivial baselines and rule-based heuristics), we will look closer at them separately.

\subsection{Trivial baselines}

As it was mentioned before, first we chose three baseline methods to solve all tasks in Russian SuperGLUE: majority class, random choice and random weighted choice. When comparing these baselines to the other methods, we should keep in mind that they do not rely on any linguistic knowledge at all. 

All three baselines show very interesting results. For the majority class baseline, the best result is the one for the RWSD task. It should be emphasized again that not only one can achieve the SOTA performance with the majority baseline here, but, at the moment of submission, half of the leaderboard models probably use this approach as their solver method, since they all have the same performance score. Simple random choice worked good on the RCB and RWSD as well. Random balanced choice outperforms the majority class approach on the DaNetQA, RWSD, TERRa, PARus, and RUSSE. 

Across all RSG tasks, the balanced random choice baseline achieves the average score of $0.385$. Language models obviously outperform this score, but the difference is marginal: only half of the systems in the leaderboard achieve a score higher than $0.5$, and the best ensemble of transformers reaches $0.679$ (the human performance is $0.811$). For some benchmarks (for example, RuCOS), the random balanced baseline \textit{outperforms} BERT and GPT-3 models. In one specific case of the RWSD benchmark, no model managed to outperform the \textit{simple majority class baseline}.

From this, we conclude that the RSG leaderboard scores should be taken with a grain of salt and compared to the trivial baselines. For example, the $0.669$ accuracy of the SOTA models on the RWSD dataset is not a sign of their `human-like comprehension abilities': it is just that these models (or their authors) could not do any better than simply predict the same label for all the instances in the test set. For other tasks, the picture is only slightly better: in most cases, the leaderboard participants managed to improve the random balanced baseline only by a small margin. Another important finding is that the class balances in the RSG test sets are similar to those in the validation and training sets: this allows one to achieve boosted accuracies by simply replicating these distributions in the test answers. This is true for all the tasks evaluated by accuracy (six of the RSG tasks). If the class labels in these six datasets were perfectly balanced, the expected average accuracy of the random baseline would be $0.472$. In the real RSG, this value is $0.538$.
This is certainly an undesired property for a test set in general; in this case it additionally makes it difficult to assess to what extent the large-scale language models' NLU performance for Russian is actually an artifact of this data leakage.

\subsection{Rule-based heuristics}

Rule-based heuristics tend to improve trivial baselines in cases of TERRa, RUSSE, RCB (considering Avg. F1 score). Here we categorize the described rules. Note that most of them are language-agnostic and can be tested on benchmarks for other languages as well. 

\begin{enumerate}
    \item Using text length (e. g. `More than 30 words in the premise'): these rules are useful for RCB, PARus, MuSeRC, TERRa, RUSSE, DaNetQA.
     \item Using binary lexical features (e. g. `Presence of \foreignlanguage{russian}{`чтобы', `будет', `от', `он')}: these rules are useful for LiDiRus, RCB, TERRa, DaNetQA.
     \item Using word forms or lemmas overlap (e. g. 'Sentences 1 and 2 use the same set of lemmas'): these rules are useful for LiDiRus, RCB, PARus, MuSeRC, TERRa, RUSSE.
     \item Other task-specific heuristics.
\end{enumerate}

The existence of such statistical cues is not a problem in itself: after all, this is what machine learning is after. What we see as problematic is the fact that the large over-parameterized models seem to mostly rely on them (judging by their performance scores which are not radically higher, and sometimes even lower than the scores of our rule-based approach). This means they do not employ valid inference strategies, and do not demonstrate anything close to much-praised `natural language comprehension'. We again emphasize that our heuristics are extremely simplistic and often boil down to counting the number of words in the sentence or to finding the lexical overlap between the question and the answer (sometimes after lemmatisation). There is no doubt that billion-parameter language models can find much more statistical cues in the training data than the authors of this paper were able to come up with. But these regularities will only work on the test instances drawn from the same general population (annotated or generated according to the same guidelines). This is \textit{pattern matching}, not \textit{language understanding}.

\section{Conclusion}
\label{sec:conclusion}
The recently introduced Russian SuperGLUE (RSG) set of natural language understanding benchmarks \cite{shavrina-etal-2020-russiansuperglue} has already attracted well-deserved attention from the Russian NLP practitioners. The RSG leaderboard is filled with the impressive performance scores produced by sophisticated language models trained with bleeding-edge deep learning architectures (BERT, GPT-3, etc) on titanic corpora of Russian. But are these scores really so impressive? In this paper, we studied what performance can be achieved for the RSG benchmarks \textit{without training any language models}. 

First, we established the performance boundaries of the trivial baselines: random choice, majority class choice and balanced random class choice (probabilities weighted by the distribution of class labels in the training data). We found that in some cases, these baselines outperform large-scale language models. Second, we moved on to find out whether the RSG datasets contain other statistical regularities. In has been shown in prior work for English and other languages that deep learning models are very prone to collecting low-hanging fruits and tracing shallow semantic and structural phenomena which help minimizing the loss on a particular dataset, instead of actually learning real linguistic generalizations.

To this end, we manually compiled a set of very simple custom rule-based heuristics for each RSG dataset (for example, `\textbf{set the label `\textit{contradiction}' if the word \foreignlanguage{russian}{не} `not' is present in the hypothesis}', etc). 
It turned out that up to 50\% and more of instances (depending on a particular dataset) might be covered by this or that heuristic. Moreover, applying these rules to actually solve the RSG (with fallback to the majority class baseline if no rule is applicable) constitutes a system which achieves a very competitive RSG average score of $0.468$. This outperforms RuGPT3-Small, on par with RuGPT3-Medium, and is very close to the BERT performance.

We conclude that most RSG datasets abound in statistical regularities which can easily be found at training time and employed at test time, without expensive and complicated language model pre-training. The reasons are arguably the same as with the English test sets\footnote{In fact, many RSG test sets are translated from English.}: compilation of benchmarks by crowd-sourcing and the natural desire of crowd-workers to fulfill the job in the easiest way possible. 

To sum up, we recommend the RSG maintainers to 1) modify the test sets to minimize the data leakage from label distributions; 2) diversify the datasets so as to eliminate at least the most striking statistical cues (it shouldn't be possible to find the correct answer by simply counting words); 3) provide official majority class and random weighted baselines. We believe this will make the Russian SuperGLUE leaderboard even more informative of the real state of the art in Russian natural language processing.

In the future, it will be useful to develop a Russian equivalent of the HANS benchmark \cite{mccoy-etal-2019-right}: a test set containing adversarial examples, or even simply examples drawn from sources substantially different from those in the RSG. It will allow to evaluate the generalization abilities of large pre-trained language models for Russian. It would also be interesting to study the correlations between the predictions of our heuristics and the predictions of the language models in the RSG leader-board, in order to find out whether they actually exploit similar rules.

Finally, in the course of working on this paper, we collected a large trove of annotation errors and generally problematic or controversial cases in the RSG datasets. We have shared these findings with the RSG maintainers, in the hope of its future improvement.

\bibliography{dialogue.bib}
\bibliographystyle{ugost2008ls}

\section*{Appendix}

\subsection{Examples for heuristics}
\label{examples}
\begin{enumerate}
    \item RCB, heuristic 1 (if the hypothesis is a sub-string of the premise, the label is likely to be entailment)

premise: \foreignlanguage{russian}{`Из материалов дела следует, что начальник одного из отделов пытался выбить субсидию заинтересованной организации за «откат»'.}(`According to the case materials, the boss of one of the departments tried to get a subsidy for the interested organisation in return for the kickback.')

hypothesis: \foreignlanguage{russian}{`Начальник одного из отделов пытался выбить субсидию заинтересованной организации за «откат»'.} (`The boss of one of the departments tried to get a subsidy for the interested organisation in return for the kickback.')

label: entailment

    \item TERRa, heuristic 1 (if the hypothesis is a sub-string of the premise, the label is likely to be entailment)
    
premise: \foreignlanguage{russian}{`«Министерство сегодня вызвало российского посла, чтобы повторить свой протест. Встречи между парламентариями – важная составляющая политических контактов», - приводятся слова главы МИД Бёрге Бренде. Он добавил, что отказ в визах вызывает сожаление, так как визит мог бы способствовать укреплению двусторонних отношений.'} (`The department summoned the Russian Ambassador to reiterate their protest. Meeting between the parliamentarians --- important part of political contacts" says Head of Department of Foreign Affairs Børge Brende. He added that the denials of visas was regretful as the visit could have helped fostering bilateral ties.')

hypothesis: \foreignlanguage{russian}{`Визит мог бы способствовать укреплению двусторонних отношений.'} (`The visit could have helped fostering bilateral ties.')

label: entailment

    \item DaNetQA, heuristic 1 (if the question starts with \foreignlanguage{russian}{`был'} (`was/were'), the label is likely to be True)

question: \foreignlanguage{russian}{Были ли в австралии аборигены?}(`Was there local tribes in Australia?')

passage: \foreignlanguage{russian}{Австралийские аборигены — коренное население Австралии, также иногда называемые «австралийскими бушменами», в языковом и расовом отношениях обособлены от других народов мира. Говорят на австралийских языках, значительная часть — только по-английски и/или на различных вариантах пиджинов. Живут, в основном, в удалённых от городов районах Северной, Северо-Западной, Северо-Восточной и Центральной Австралии, часть — в городах. Австралийская цивилизация является одной из старейших непрерывных культур в мире. В расовом отношении аборигены Австралии образуют отдельную — собственно австралийскую ветвь австралоидной расы.} (`Aboriginal Australians are the indigenous people of Australia, who are sometimes called Indigenous Australians as well. In terms of language and their race they differ significantly from other world's peoples. They speak Australian language, yet the majority can speak only English or its various plain versions. Most of them live in the outlying areas of North, North-Western or Central Australia, though some of them live in the cities. Australian civilisation is one of the oldest and fundamental world's cultures. In terms of race Aboriginal Australians form a separate Australian-only branch of Australo-Melanesian race.')

label: True

    \item DaNetQA, heuristic 4 (if the question starts with \foreignlanguage{russian}{`едят ли'} (`do they eat'), the label is likely to be False)

question: \foreignlanguage{russian}{Едят ли в греции греческий салат?} (`Do Greeks eat Greek salad?')

passage: \foreignlanguage{russian}{Греческий салат — греческий салат из помидоров, огурцов, феты, шалота и маслин, заправленный оливковым маслом с солью, чёрным перцем, орегано. Ключевым компонентом салата является фета — традиционный греческий сыр из овечьего или козьего молока. Часто в салат добавляют сладкий перец, реже — каперсы или анчоусы. В англоязычных странах в рецепт всегда включают листовой салат, обычно — лук или сладкий перец; иногда добавляют и другие ингредиенты. В самой Греции такие варианты почти не встречаются.}(`Greek salad is a salad that consists of tomatoes, cucumbers, feta cheese, shallots and olives with Olive oil, sail, black pepper and oregano. The key ingredient is feta --- traditional Greek cheese made of sheep or goat’s milk. It often happens that sweet peppers are added to the salad, rarely --- capers or anchovies. In English-speaking countries the recipe always includes lettuce or kale, often onion or bell pepper. Sometimes other ingredients are added. In Greece itself such variants are rarely seen.')

label: False

\item TERRa, heuristic 8 (the presence of \foreignlanguage{russian}{`только', `мужчина'} (`only', `man') leads to not\_entailment)

premise: \foreignlanguage{russian}{`"Была установлена личность подозреваемого - 27-летнего мужчины. По словам задержанного, он был давно влюблен в жену убитого и различными способами добивался ее внимания. ""Так как женщина не хотела с ним общаться, он решил похитить ее мужа"", - говорится в сообщении."'} (`The suspect was identified as 27 year old man. According to the apprehended, he had long been in love with the killed man's wife and tried hard to win her over. Since the woman did not want to have anything to do with him, he had decided to kidnap her husband' says the note.)

hypothesis: \foreignlanguage{russian}{`27-летний мужчина похищен.'} (`27 year old man was kidnapped.')

label: not\_entailment

\item TERRa, heuristic 8 (the presence of \foreignlanguage{russian}{`только', `мужчина'} (`only', `man') leads to not\_entailment)

premise: \foreignlanguage{russian}{`Недавно стало известно, что в общественном транспорте столицы и областных городов подорожает проезд.'} (`Recently it became known that the fare in the capital and and provincial centers is going to rise.')
hypothesis: \foreignlanguage{russian}{Проезд подорожает только в общественном транспорте столицы.} (`The fare is going to rise only in the capital.')

label: not\_entailment

\item DaNetQA, heuristic 3 (if the question starts with \foreignlanguage{russian}{`входит ли'} (`does it belong to'), the label is likely to be False)

question: \foreignlanguage{russian}{`Входит ли финляндия в скандинавию?'} (`Is Finland part of Scandinavia?')

passage: \foreignlanguage{russian}{(`Страны Северной Европы — культурно-политико-географический регион в Северной Европе и Северной Атлантике, включающий в себя государства Скандинавии — Данию , Швецию и Норвегию — и исторически связанные с ними государства Финляндию и Исландию. Иногда все эти государства называют скандинавскими странами или Скандинавией, что не совсем корректно. Понятие страны Северной Европы включает в себя понятие Скандинавия, но гораздо шире последнего. В понятие Скандинавия, как правило, не включают территории, находящиеся за пределами Европы , острова, находящиеся на большом удалении от Скандинавского полуострова, и Финляндию. Страны Северной Европы расположены в северо-западной части Европы и на островах северной Атлантики и Северного Ледовитого океана.')} (`The Nordic countries are a cultural-political-geographical region in Northern Europe and the North Atlantic, which includes the Scandinavian states - Denmark, Sweden and Norway - and the historically connected states of Finland and Iceland. Sometimes all these states are called the Scandinavian countries or Scandinavia, which is not entirely correct. The concept of the Nordic country includes the notion of Scandinavia, but the former is broader than the latter. Usually the notion of Scandinavia does not include territories outside Europe, islands located at a great distance from the Scandinavian peninsula, and Finland. The Nordic countries are located to the northwest of Europe and on the islands of the North Atlantic and the Arctic Ocean.')

label: False

\end{enumerate}

\end{document}